\documentclass{article}

\PassOptionsToPackage{numbers}{natbib}

\usepackage[final]{neurips_2018}

\usepackage[utf8]{inputenc} 
\usepackage[T1]{fontenc}    
\usepackage{hyperref}       
\usepackage{amsmath}
\usepackage{url}  
\usepackage{graphicx}
\usepackage{wrapfig}
\usepackage{algorithm}
\usepackage{adjustbox}
\usepackage{subcaption}
\usepackage{algorithmic}
\usepackage{booktabs}       
\usepackage{amsfonts}       
\usepackage{nicefrac}       
\usepackage{microtype}      
\usepackage{enumitem}

\newcommand{\bs}[1]{\boldsymbol{#1}}

\title{A Bandit Approach to Posterior Dialog Orchestration Under a Budget}

\author{
  Sohini Upadhyay, Mayank Agarwal, Djallel Bounneffouf, and Yasaman Khazaeni \\
  IBM Research AI\\
  \texttt{\{firstname.lastname\}@ibm.com} \\
}

\begin{document}

\maketitle

\begin{abstract}

Building multi-domain AI agents is a challenging task and an open problem in the area of AI. Within the domain of dialog, the ability to orchestrate multiple independently trained dialog agents, or skills, to create a unified system is of particular significance. In this work, we study the task of online posterior dialog orchestration, where we define posterior orchestration as the task of selecting a subset of skills which most appropriately answer a user input using features extracted from both the user input and the individual skills. To account for the various costs associated with extracting skill features, we consider online posterior orchestration under a skill execution budget. We formalize this setting as Context Attentive Bandit with Observations (CABO), a variant of context attentive bandits, and evaluate it on simulated non-conversational and proprietary conversational datasets. 



\end{abstract}

\section{Introduction}

Serverless execution enables scalable and modular deployment of models for contemporary applications, including AI agents. In the context of dialog systems, such modular design entails connecting multiple dialogue agents or "skills" - each trained independently on different or overlapping tasks - to form a unified system with capabilities of each of its constituent skills. The orchestrator or control module for such a system can either be a deterministic module employing “If this then that” (IFTTT) logic or more complex functional programming frameworks such as Amazon Lambda etc, or could in itself be a learnable model employing either supervised or reinforcement learning approaches. 

This is a fairly common scenario in contemporary personal home assistant devices such as Amazon Alexa and Google Home, where developers have the ability to integrate their own independently developed skills with the assistant's core infrastructure. Here, the assistant itself is responsible for invoking skills in response to user input. Invocation of these skills falls into two categories: explicit invocation and implicit invocation. Explicit invocation occurs when the user explicitly specifies the name of the skill they are interested in interacting with. This requires the user to specify the name of the skill along with a set of pre-defined invocation phrases that trigger the skill \cite{googleInvocation}, and is an application of the IFTTT logic. Implicit invocation on the other hand, does not provide the assistant with the name of the skill the user is interested in interacting with, and requires the assistant to understand the users query along with the available skills capabilities to select the most appropriate skill to respond with \cite{googleInvocation, amazonInvocation}. Implicit invocation has a clear advantage in facilitating more natural conversation and removes the knowledge barrier of skill naming and understanding that explicit invocation requires. 

Dialog orchestration models that use implicit invocation tend to follow the a-priori approach or the posterior approach. A-priori orchestrator models are built exclusively using features known prior to executing any skills whereas posterior models execute skills to extract supplemental features. By this definition, it is clear that posterior approaches ought to match or beat comparable a-priori methods as they use a superset of the features used in the a-priori approach. Most recent work has focused on the posterior approach, including submissions to the Alexa prize competition \cite{Papaioannou2017,Adewale2017}. Both of these approaches have leveraged supervised learning techniques, which necessitate training data and regular updating if deployed live.  

Online orchestration models could remove this hurdle, enabling a cold start deployment. Online orchestration also does not require a fixed label space, allowing new skills to be added to the agent in a seamless way. While a multitude of reinforcement learning models are viable orchestration candidates, we investigate the use of contextual bandits for the task. The contextual bandit problem is a variant of the extensively studied multi-armed bandit problem ~\cite{LR85,gittins1979bandit,UCB}, where at each iteration, before choosing an arm, the agent observes an $N$-dimensional {\em context}, or {\em feature vector}, and uses it to predict the next best arm to play \cite{langford2008epoch,agarwal2009explore,auer2002nonstochastic,AgrawalG13}.
Every time an arm is played, a reward value is observed. Over time, the agent's aim is to collect enough information about the relationship between the context vectors and rewards, so that it can predict the next best arm to play by looking at the corresponding context \cite{langford2008epoch,AgrawalG13}. 

Posterior orchestration, online or otherwise, is not without its challenges. Recall that in posterior dialog orchestration, a user's query is often directed to a number of domain specific skills and the best response is returned. In this case, the pre-execution features, i.e features extracted from the query, can be immediately observed, but the set of features or responses from skills, the post-execution features, cannot. For multi-purpose dialog systems, like personal home assistants, executing and retrieving features or responses from every skill can be computationally expensive or intractable, with the potential to cause a poor user experience. Moreover, executing skills in some use cases may necessitate api requests associated with actual costs. Thus while posterior dialog orchestration models are in many ways conceptually preferable to a-priori approaches, in practice they are associated with an often unaccounted cost. The challenge here is introducing a budget on the number of post-execution features that can be extracted. Some existing supervised posterior orchestration methods recognize this challenge and avoid retrieving all post-execution features. As an example, Kim et. al. \cite{Kim2018} present a set of efficient and scalable neural shortlisting-reranking models for personal assistants. The  shortlisting stage efficiently  trims  all  the skills down to a list of top-k candidates, and  the  reranking  stage  performs  a  list-wise reranking of the initial top-k skills with additional contextual information. Beyond the lack of cold-start support inherent to all supervised approaches, the amount of data necessary to effectively train this kind of neural model based method is a limiting factor for low-data use cases.
Given that online orchestration avoids these pitfalls, we develop a novel bandit algorithm that handles this challenge of limited access to post-execution features. 

The goal of our research is to build a dialog orchestration framework which can utilize query and user features along with the conversational context to route the dialog in a multi-skill system. Overall, the main contributions of this paper include
(1) presenting an online approach to dialog orchestration,
(2) a new variant of the context attentive bandit problem, motivated by limitations of posterior dialog orchestration, and (3) an empirical evaluation demonstrating the advantages of our proposed method over a range of datasets and settings.

\section{Background}
The contextual bandit problem has been extensively studied in the past, and a variety of solutions have been proposed. In LINUCB ~\cite{li2010contextual,abbasi2011improved,chu2011contextual} and in Contextual Thompson Sampling (CTS)~\cite{AgrawalG13}, a linear dependency is assumed between the expected reward given the context and an action is taken after observing this context; the representation space is modeled using a set of linear predictors. However, the context is assumed to be {\em fully observable}, which is not the case in this work.  

Motivated by dimensionality reduction tasks, Abbasi-Yadkori et. al. \cite{YadkoriPS12} studied a sparse variant of stochastic linear bandits, where only a relatively small and unknown subset of features is relevant to a multivariate function optimization. Similarly, Carpentier \& Munos \cite{CarpentierM12} also considered  high-dimensional stochastic linear bandits with sparsity, 
where $s$ components are assumed to be non-zero, and where the dimension $N$ of the context is larger than the sampling budget $n$. 
In Bastani \& Bayati \cite{bastani2015online} consider a multi-arm bandit (MAB) problem with high-dimensional covariates, and a new efficient bandit algorithm based on the LASSO estimator is presented. Regret analysis is performed, demonstrating that the proposed algorithm achieves near-optimal performance in comparison to an oracle that knows all the problem parameters. Still, all above work, unlike ours, assumes  full observability of the context variables, which is not the case in many important applications.

Finally, Bouneffouf et. al. \cite{BouneffoufRCF17} developed the idea of context attentive bandits - a case of the contextual bandit problem, referred to as contextual bandit with restricted context (CBRC), where observing the  whole feature vector at each iteration is impossible, and the agent can only request to see some limited number of those features; the upper bound (budget) on the feature subset is fixed for all iterations, but within this budget, the agent can choose any feature subset of said size. However, in the posterior dialog orchestration application, while the full context may be too costly or impossible to see, some partial observation of the context, e.g. query or user features, can be known to an agent initially, along with the ability to observe unknown context features, up to a certain limit, as in CBRC.

Motivated by the limitations of posterior dialog orchestration, we extend the context attentive bandit to a special case which we call the {\em Context  Attentive Bandit with Observations (CABO)}. In the CABO setting, observing the full context vector at each iteration is impossible, but a small subset of  context features, is observable and a fixed number of the unobserved features within a budget can be revealed. The goal here is to leverage the observable features to select the best unknown feature subset at each iteration to maximize overall reward. 

\section{Problem Setting}
We begin by formally defining concepts our novel bandit problem setting builds upon, such as contextual bandit and contextual combinatorial bandit.

\emph{\bf{The Contextual Bandit Problem.}}
Following Langford \& Zhang \cite{langford2008epoch}, this problem is defined as follows.
At each time point (iteration) $t \in \{1,...,T\}$, an agent is presented with a {\em context} (feature vector) $\textbf{c}(t) \in \mathbf{R}^N$
  before choosing an arm $k  \in A = \{ 1,...,K\} $.
We denote by
  $C=\{C_1,...,C_N\}$ the set of features (variables) defining the context.
Let ${\bf r(t)} = (r_{1}(t),...,$ $r_{K}(t))$ denote  a reward vector, where $r_k(t) \in [0,1]$ is a reward at time $t$  associated with the arm $k\in A$.
Herein, we will primarily focus on the Bernoulli bandit with binary reward, i.e. $r_k(t) \in \{0,1\}$.
Let $\pi: \mathbf{R}^N \rightarrow A$ denote a policy,
mapping a context  $\bs{c}(t) \in \bs{R}^N$ into an action $k \in A$.  We assume some probability distribution $P_c(\bs{c})$ over the contexts in $C$, and a distribution  of the reward, given the context and the action taken in that context.
We assume that the expected reward (with respect to the distribution $P_r(r|\bs{c},k)$) is a  linear function of the context, i.e.
$E[r_k(t)|\textbf{c}(t)] $ $= \mu_k^T \textbf{c}(t)$,
where $\mu_k$ is an unknown weight vector associated with the arm $k$; the agent's objective is to learn $\mu_k$  from the data so it can optimize its cumulative reward over time.   \\

\emph{\bf {Contextual Combinatorial Bandit.}}
Our feature subset selection approach builds upon the {\em Contextual Combinatorial Bandit (CCB)} problem \cite{qin2014contextual}, specified as follows.
Each arm $k \in \{1,...,K\}$ is associated with the corresponding variable $x_{k}(t)\in R$ indicating the reward obtained when choosing the $k$-th arm at time $t$, for $t>1$. 
In the contextual combinatorial bandit setting, the agent sequentially observes a context $\bs{c}$, selects a subset of arms $M \in S$, from a constrained set of arm subsets $S \subseteq P(K)$, where $P(K)$ is the power-set of $K$, and observes a reward $r_{M}(t) = h(x_{k}(t)), k \in M$ associated with the selected subset of arms. 
Here we define the reward function $h(\cdot)$ used to compute $r_{M}(t)$ as a sum of the outcomes of the arms in $M$, i.e.   $r_{M}(t)=\sum_{k\in M} x_{k}(t)$, although one can also use nonlinear rewards. The objective of the CCB algorithm is to maximize the reward over time. We consider here a stochastic model, where  the expectation of $x_i(t)$ observed for an arm $k$ is a linear function of the context, i.e.
$E[x_i(t)|\textbf{c}(t)] $ $= \mu_i^T \textbf{c}(t)$,
where $\mu_i$ is an unknown weight vector (to be learned from the data) associated with the arm $i$. The distributions can be different for each arm. The global rewards $r_{M}(t)$ are also random variables, independent and distributed according to some unknown distribution with some expectation $\mu^M$.


\subsection{CABO: Context Attentive Bandit with Observations}
Building off the contextual bandit and contextual combinatorial bandit problems, we formally define a novel type  of  bandit problem,  called  {\em Context Attentive Bandit with Observations (CABO)}.

As mentioned above,  $\textbf{c}(t) \in \mathbf{R}^N$ will denote a vector of values  assigned to an (ordered) set of random   context variables, or features, $C = \{C_1,...,C_N\}$, at time $t$. 
Let $ C^D \subseteq C$, $|C^D|=D$, $0 < D \leq N$, denote a subset of features of size $D$, and  
let $\textbf{c}^D(t) \in S_{C^D}$ denote a vector from a $D$-subspace of $\mathbf{R}^N$, denoted  $S_{C^D} \subseteq \mathbf{R}^N$, which is defined as a subspace containing  all sparse vectors with   features (coordinates) outside of the  subset  $C^D$ set to zero.


We assume that at each time point $t$ the environment generates a feature vector $\mathbf{c}(t) \in \mathbb{R}^N$ which the agent cannot  observe fully. However,  unlike the previously introduced CBRC setting  \cite{BouneffoufRCF17}, the agent has now a {\em partial observation} of the context, i.e. it can see a small subset of {\em observed}  features $C^O \subset C$, where $O << N$. Given these observations $\bs{c}^O$, the agent is allowed to request more features to observe (similar to  CBRC setting),   up to   $D$ (desired) features in total, including the initial set $C^O \subset C^D$, where $C^D$ denotes  final set of $D$ observed features. Assuming that the unobserved features are all of the same fixed cost, there is a budget of $U=D-O$ features imposed on the agent. The goal of the agent  is   to maximize its total reward over time via (1) the optimal choice of the additional  observations, given the initial ones,   and (2) the optimal choice of a subsequent action $k \in A$ based on the resulting  extended  observation.  

Let us now formally define the set of all policies, i.e.  possible mappings from agent's observations to its actions restricted to the proposed  problem setting, as the set of the compound functions  
\begin{eqnarray}
\nonumber
\Pi^D_o = \cup_{C^O \subseteq C} \{\pi: S_{C^O} \rightarrow A, s.t. ~ \pi(\bs{c}^O)=\hat{\pi}(\bs{c}^D), \\     \bs{c}^D = g(C^D), C^D = h(C^O,\bs{c}^O) \},
\label{policy}
\end{eqnarray}
where 
\begin{itemize}
\setlength{\itemindent}{-.25in}
\item $~g: P_D(N) \rightarrow S_D(\mathbf{R}^N)$ is a function mapping a given subset of features $C^D \in P_D(N)$, $P_D(N)$ denoting the set of all subsets of $\{1,...,N\}$ of size $D$, to a vector $\bs{c}^D \in S_D(\mathbf{R}^N)$, $S_D(\mathbf{R}^N)$ denoting the set of all $d$-subspaces $S_{C^D}$ of $\mathbf{R}^N$, each defined for a corresponding subset $C^D$ of features; 
\item $~h: P_O(N) \rightarrow P_D(N)$  maps the initial set of observed features $C^O$ to the extended set of features to be observed, $C^D$, $C^O \subset C^D$; 
\item $~\hat{\pi}: S_{C^D} \rightarrow A$ is a function mapping the observed extended feature subset $\bs{c}^D$ into an action (a.k.a. bandit's arm) $k(t) \in A$, which results into a reward $r_{k}(t)$. 
\end{itemize}
The objective of a contextual bandit algorithm is to find an optimal  policy $\pi \in \Pi^D_O$, over $T$ iterations or time points,  so that the total reward is maximized. 

\section{Methodology}
\subsection{CATSO: Context  Attentive Thompson Sampling with Observations}
\begin{scriptsize}
\begin{algorithm}
 \caption{Context Attentive Thompson Sampling with Observations}
\label{alg:CATSO}
\begin{algorithmic}[1]
 \STATE {\bfseries }\textbf{Require:} Total number of features $N$; initially observed number of features $O$;  the set of those features and their values, $C^O$ and $\bs{c}^O$, respectively;  the total desired number of features to observe $D$, over $S$ stages; the  exploration parameter $v$, and the function $\lambda(t)$ computed differently for stationary and nonstationary cases.
 \STATE {\bfseries }\textbf{Initialize:} $\forall k \in \{1,...,K\}, A_k=I_K$, $g_k = 0_K$, $\hat{\mu_k}= 0_K$, and $\forall i \in \{1,...,N\}$, $B_i=I_N$, $z_i=0_N$, $\hat{\theta_i}=0_N$.
 \STATE {\bfseries }$U = D-O$, $~u = U/S$ 
 \STATE {\bfseries }\textbf{Foreach} $t \in 1, 2, . . . ,T$ \textbf{do}
 \STATE \quad observe  $\bs{c}^O(t)$, given feature subset $C^O$  
 \STATE \quad $X^t = C^O$, $\bs{x}(t) = \bs{c}^O$
 \STATE {\bfseries }\quad\textbf{Foreach} stage $s = 1, 2, . . . ,S$ \textbf{do}
 \STATE {\bfseries }\quad \quad \textbf{Foreach} context feature $i= 1,...,N$ \textbf{do}
 \STATE \quad \quad \quad \textbf{if} $i \notin X^t$ \textbf{then}
 \STATE {\bfseries }\quad \quad\quad\quad Sample $\theta_i$  from $\mathcal{N}(\hat{\theta}_i, v^2 B_i^{-1})$
 \STATE \quad \quad \quad \textbf{End if}
\STATE {\bfseries } \quad\quad\textbf{End do}
\STATE \quad\quad Select $C^{u}(t)=\underset{C' \subseteq C/X^t, |C'|=u}{\text{argmax}} \sum_{i \in C' } \bs{x}(t)^\top \theta_i$
\STATE \quad \quad $X^{t} = X^{t} +C^{u}(t)$
 \STATE  {\bfseries }\quad\quad
observe values  $\bs{x}(t)$ of feature subset $X^t$ 
\STATE \quad \textbf{End do}
 \STATE {\bfseries }\quad\textbf{Foreach} arm $k= 1,...,K$ \textbf{do}
 \STATE {\bfseries }\quad \quad Sample $\mu_k$ from $\mathcal{N}(\hat{\mu}_k, v^2 A_k^{-1})$ distribution.
 \STATE {\bfseries } \quad\textbf{End do}
 \STATE {\bfseries }\quad Select arm $k(t)= \underset{k\subset \{1,...,K\} }{\text{argmax}} \ \bs{x}(t)^\top \mu_k$
 \STATE {\bfseries }\quad Observe $r_{k}(t)$
\STATE \quad$A_k= A_{k}+ \bs{x}(t)\bs{x}(t)^\top $
\STATE \quad $g_k = g_k + \bs{x}(t)r_{k}(t)$
\STATE \quad $\hat{\mu}_k = A_k^{-1} g_k$
\STATE \quad \textbf{Foreach} $i \in X^t  \setminus C^O$
\STATE \quad\quad$B_i= \lambda(t) B_{i}+ \bs{x}(t)\bs{x}(t)^{\top} $
\STATE \quad\quad $z_i = z_i + \bs{x}(t)r_{k}(t)$
\STATE \quad\quad$\hat{\theta}_i = \lambda(t) B_i^{-1} z_i$
 \STATE {\bfseries }\quad \textbf{End do}
 \STATE {\bfseries }\textbf{End do}
\end{algorithmic}
\end{algorithm}
\end{scriptsize}
We propose a novel method for solving the CABO problem, which we name {\em Context  Attentive Thompson Sampling with Observations (CATSO)}, and summarize it in Algorithm \ref{alg:CATSO}.
The combinatorial task of selecting the best subset of features is treated as a contextual combinatorial bandit (CCB) problem \cite{qin2014contextual}, and the subsequent decision-making (action selection) task as a contextual bandit problem solved by Contextual Thompson Sampling (CTS) \cite{AgrawalG13}, respectively.

The algorithm takes the total number of features $N$, initially observed number of features  $O$, and the total desired  number of features to observe $D$, as inputs. We use $U=D-O$ to denote our budget, the number of unobserved features to reveal. We will use several stages, up to $S$, to reveal $U$ features. When $S=1$, the observed features $O$ are used to select all $U$ features as a set, whereas when $S=U$, the set of $O$ features is updated incrementally and used to select each of the $U$ additional features one at a time.  The algorithm also requires hyperparameter $v$, the exploration parameter used in Thompson Sampling. 

The algorithm iterates over $T$ steps, where at each iteration $t$, the values $\bs{c}^O(t)$ of features in the original observed subset $C^O$ are observed first. The current set of already observed features, $X^t$, and the corresponding observed context, $\bs{x}(t)$, is maintained over all stages, and are initialized to $C^O$ and $\bs{c}^O$ respectively.
At each iteration $t$,  the vector parameter $\theta_i$ is sampled from the corresponding multivariate Gaussian distribution (step 10) for each feature $i$ not yet observed so far, to estimate $\hat{\theta_i}$.
Thereafter, at each stage, the best subset of features are selected, $C^u \subseteq C/X^t$, such that $C^u=\arg \max_ {C' \subseteq C/X^t, |C'|=u} \sum_{i \in C'} {\bs{x}(t)}^\top \theta_i $ where $u =U/S$ is the number of unknown features to explore at each stage.

Once a subset of features is selected using the contextual combinatorial bandit approach, the algorithm switches to the contextual bandit setting to choose an arm based on the context consisting now of a subset of features (steps 17-24).

We assume that the expected reward is a linear function of a restricted context, 
$$E[r_k(t)|\textbf x(t)]= \mu_k^T \textbf x(t)$$ We assume
that  reward $r_{k}(t)$ for choosing arm $k$ at time $t$   follows a parametric likelihood function $P(r(t)|\mu_k)$, and that
the posterior distribution at time $t + 1$, $P(\mu|r(t)) \propto P(r(t)|\mu) P(\mu)$, is given by a multivariate Gaussian distribution $\mathcal{N}(\hat{\mu_k}(t+1)$, $v^2 A_k(t + 1)^{-1})$ where
$A_k(t)= I_N + \sum^{t-1}_{\tau=1} c(\tau) c(\tau)^\top$
with $N$ the size of the context vectors $c$, 
and $\hat{\mu_k}=A_k(t)^{-1} (\sum^{t-1}_{\tau=1} c(\tau) c(\tau))$.

At each time point $t$, and for each arm, a $k$-dimensional $\mu_k$ is sampled from 
$\mathcal{N}(\hat{\mu_k}(t)$, $ v^2{A_k(t)}^{-1})$ , an arm is chosen such that $\textbf{x}(t)^\top \mu_k$ is maximized (step 20 in the algorithm), a reward $r_k(t)$ is obtained for choosing an arm $k$, and finally the relevant parameters are updated.

\subsubsection{CATSO in Nonstationary Setting}
Practical posterior dialog orchestration applications motivate the need to consider the possibility of nonstationary unobserved context features. In posterior dialog orchestration, we assume each domain specific skill outputs features pertaining to their query response. In some use cases, each skill could be independently updated at any time, changing these features. As a result, similar queries, which would likely define the observable context $C^O$, can elicit vastly different distributions of response features, the unknown context, over time. The main problem with any stationary algorithm is that it gives equal weight to its history. In a nonstationary environment, if there is no specific assumption about how the environment will change, a simple idea is to
use a weighting function to lessen the effect of the past on current decisions. Since we are using CTS as our base model and it uses ridge regression, implementation of weighting instances is straightforward. We propose assigning decaying weights to the past examples in the ridge regression. The same kind of weights are also applied in the calculation of the confidence width. 
Following the notation from Algorithm \ref{alg:CATSO}, in this case $\lambda(t)$ represents the decay parameter. In order to compute the optimal $\lambda(t)$ value, we use GP-UCB algorithm \cite{srinivas2009gaussian}, which is an algorithm that solves the  multi-armed bandit problem in continuous space. Computing $\lambda(t)$ is done via the following decision rule:
\begin{eqnarray*}
\lambda(t)= argmax_{\lambda \in D}[\mu_{t-1}(\lambda)+\alpha^{1/2} \sigma_{t-1}(\lambda)]
\end{eqnarray*}
 with $\alpha$ as the GP-UCB exploration parameter, $\mu_{t-1}$ mean reward, and $\sigma_{t-1}$ the uncertainty. The GP-UCB algorithm is initialized with the search space $ \lambda \in [0,1]$ and at each iteration uses the above equation to calculate a different $\lambda(t)$, which is then used by CATSO. More detail on the GP-UCB algorithm can be found in \cite{srinivas2009gaussian}.

\section{Experiments}

We assess Context Attentive Thompson Sampling with Observations (CATSO) with respect to the current state of the art for context attentive bandits, Thompson Sampling with Restricted Context (TSRC).  TSRC solves the contextual bandit with restricted context problem (CBRC) discussed prior, which selects a set of unknown features at each event while assuming no observable features exist initially. For a total number of features $N$, we refer to the $O$ observed features as the known context and the $N-O$ unobserved context features as the unknown context. In our use of the TSRC algorithm, at each iteration, the known context is observed, the TSRC decision mechanism independently chooses $U$ unknown context features to reveal, and Contextual Thompson Sampling (CTS) is invoked. Empirical evaluation of CATSO and TSRC was performed on publicly available classification datasets and on a propriety corporate dialog orchestration dataset. 

Publicly available Covertype\footnote{https://archive.ics.uci.edu/ml/datasets.html} and CNAE-9$^1$ were featured in the original TSRC paper and Warfarin \cite{sharabiani2015revisiting} is a historically popular dataset for evaluating bandit methods. The details of these datasets are summarized in Table \ref{table:Synthetic}. 

\begin{wraptable}{r}{5.5cm}
	\centering
	\vspace{-10pt}
	\caption{Datasets}
	\resizebox{\linewidth}{!}{
		\begin{tabular}{ l | c | r | r }
			Datasets                  & Instances    & Features   & Classes \\ \hline
            Covertype                & 500 000      & 95           &  7\\
            CNAE-9                   & 1080         & 856          &  9\\
            Warfarin				 & 5528         & 93           & 3 
			
		\end{tabular}
	}
	\label{table:Synthetic}
\end{wraptable}

For the stationary setting, we randomly fix $10\%$ of the context feature space of each dataset to be known at the onset and explore a subset of $U$ unknown features. For CATSO, we fix $\lambda(t) = 1$ to reflect the stationary setting and choose $S=1$. For the nonstationary setting, we simulate nonstationarity in the unknown feature space by duplicating each dataset, randomly fixing the known context in the same manner as above, and shuffling the unknown feature set - label pairs. Then we stochastically replace events in the original dataset with their shuffled counterparts, with the probability of replacement increasing uniformly with each additional event. For this nonstationary setting, which we refer to as NCATSO, we again fix $S=1$, but use $\lambda(t)$ as defined by the GP-UCB algorithm. We compare NCATSO to Weighted TSRC (WTSRC), the nonstationary version of TSRC also developed by  Bouneffouf et al. \cite{BouneffoufRCF17}. WTSRC makes updates to its feature selection model based only on recent events, where recent events are defined by a time period, or "window" $w$. We choose $w=100$ for WTSRC. We report the total average reward across a range of $U$ corresponding to various percentages of $N$ for each algorithm in each setting in Table \ref{tab:uci}. 
\begin{table}[h]
\caption{Total average reward, $O=10\%$}
\label{tab:uci}
\begin{subtable}[h]{0.49\linewidth}\centering
\caption{Stationary setting}
\label{table:stationary}
\begin{adjustbox}{width=\linewidth}
\begin{tabular}{c|l|l|l}  
 & \multicolumn{3}{c}{\textbf{\textit{Warfarin}}}\\
\hline
{\textit{U} }      & 20\%                        & 40\%                        & 60\%                        \\ \hline
 {TSRC}         & 53.28 $\pm$ 1.08          & 57.60 $\pm$ 1.16          & 59.87 $\pm$ 0.69                \\ \hline
{CATSO}       & \textbf{53.65 $\pm$ 1.21} & \textbf{58.55 $\pm$ 0.67} & \textbf{60.40 $\pm$ 0.74}         \\ \hline
\end{tabular}
\end{adjustbox}
\begin{adjustbox}{width=\linewidth}
\begin{tabular}{c|l|l|l} 
  & \multicolumn{3}{c}{\textbf{\textit{Covertype}}} \\ \hline
{\textit{U }}              & 20\%                        & 40\%                        & 60\%             \\ \hline
TSRC               & 54.64 $\pm$ 1.87          & 63.35 $\pm$ 1.87          & 69.59 $\pm$ 1.72           \\ \hline
CATSO              & \textbf{65.57 $\pm$ 2.17} & \textbf{72.58 $\pm$ 2.36} & \textbf{78.58 $\pm$ 2.35}\\ \hline
\end{tabular}
\end{adjustbox}
\begin{adjustbox}{width=\linewidth}
\begin{tabular}{c|l|l|l}
                    & \multicolumn{3}{c}{\textbf{\textit{CNAE-9}}}                                                                                                                           \\ \hline
{\textit{U }}  & 20\%                        & 40\%                        & 60\%                                 \\ \hline
TSRC               & \textbf{33.57 $\pm$ 2.43} & 38.62 $\pm$ 1.68          & \textbf{42.05 $\pm$ 2.14}         \\ \hline
CATSO             & 29.84 $\pm$ 1.82          & \textbf{39.10 $\pm$ 1.41} &40.52 $\pm$ 1.42 \\ \hline
\end{tabular}
\end{adjustbox}
\end{subtable}
\hfill
\begin{subtable}[h]{0.49\linewidth}\centering
\centering
\caption{Nonstationary setting}
\label{table:nonstationary}
\begin{adjustbox}{width=\linewidth}
\begin{tabular}{c|l|l|l}
                    & \multicolumn{3}{c}{\textbf{\textit{Warfarin}}}                                                                                                                       \\ \hline
\textit{U} & 20\%                        & 40\%                        & 60\%                     \\ \hline
WTSRC               & 55.83 $\pm$ 0.55          & 58.00 $\pm$ 0.83          & 59.85 $\pm$ 0.60          \\ \hline
NCATSO              & \textbf{59.47 $\pm$ 2.89} & \textbf{59.34 $\pm$ 2.04} & \textbf{63.26 $\pm$ 0.75} \\ \hline
\end{tabular}
\end{adjustbox}
\begin{adjustbox}{width=\linewidth}
\begin{tabular}{c|l|l|l}
                    & \multicolumn{3}{c}{\textbf{\textit{Covertype}}   }                                                                                                                   \\ \hline
\textit{U}  & 20\%                        & 40\%                        & 60\%                        \\ \hline
WTSRC               & \textbf{50.26 $\pm$ 1.58} & 58.99 $\pm$ 1.81          & 64.91 $\pm$  1.38         \\ \hline
NCATSO              & 48.50 $\pm$ 1.05          & \textbf{68.17 $\pm$ 3.14} & \textbf{83.78 $\pm$ 5.51} \\ \hline
\end{tabular}
\end{adjustbox}

\begin{adjustbox}{width=\linewidth}
\begin{tabular}{c|l|l|l}
                    & \multicolumn{3}{c}{\textbf{\textit{CNAE-9}}}                                                                                                                           \\ \hline
\textit{U}  & 20\%                        & 40\%                        & 60\%                   \\ \hline
WTSRC               & 19.91 $\pm$ 2.67          & 30.86 $\pm$ 2.92          & 36.01 $\pm$ 2.88       \\ \hline
NCATSO              & \textbf{30.88 $\pm$ 0.96} & \textbf{34.91 $\pm$ 1.93} & \textbf{42.04 $\pm$ 1.52} \\ \hline
\end{tabular}
\end{adjustbox}
\end{subtable}
\end{table}
The results in Table \ref{tab:uci} are promising, with our methodologies outperforming the state of the art in the majority of cases across both settings. The most notable exception is where CATSO sometimes outperforms and other times nearly matches TSRC performance on the CNAE-9 dataset. This outcome is somewhat expected, for in the original work on TSRC \cite{BouneffoufRCF17}, the mean error rate of TSRC was only $0.03\%$ lower than randomly fixing a subset of unknown features to reveal for each event on CNAE-9. This suggests that the operating premise of TSRC, that some features are more predictive of reward than others, does not hold on this dataset. On top of this assumption, CATSO also assumes that there exist relationships between the known and unknown context features, likely causing a small compounding of error. 

Proprietary corporate dialog application \textit{Customer Assistant} orchestrates 9 domain specific skills which we arbitrarily denote as $Skill_{1}, \ldots,  Skill_{9}$ in the discussion that follows. In this application, example skills lie in the domains of payroll, compensation, travel, health benefits, and so on. Each skill is designed with a multi-turn conversation dialog tree. In addition to a textual response to a user query, the skills orchestrated by \textit{Customer Assistant} also return the following features: an \textit{intent}, a short string descriptor that categorizes the perceived intent of the query, and a \textit{confidence}, a real value between 0 and 1 indicating how confident a skill is that its response is relevant to the query. Skills have multiple intents associated with them. The orchestrator uses all the features associated with the query and the candidate responses from all the skills to choose which skill should carry the conversation at a given event.

We accessed the training data for each skill to find example queries \textit{Customer Assistant} has valid responses for, amounting to 28,412 queries total. Accordingly, we denote the correct class for a query to be the skill it was an example for. For 127 queries common to more than one skill's training data, one of the skills was randomly assigned as the correct class. We pose each query to all of the skills to extract the associated intents and confidences, and add noise sampled from a $\mathcal{N}$(-0.1, 0.05) distribution to all of the confidences to avoid built-in biases. We encode each query by averaging 50 dimensional GloVe word embeddings \cite{pennington2014glove} for each word in each query and for each skill we create a feature set consisting of its confidence and a one-hot encoding of its intent. The skill feature set size for $Skill_{1}, \ldots,  Skill_{9}$ are 181, 9, 4, 7, 6, 27, 110, 297, and 30 respectively. We concatenate the query features and all of the skill features to form a 721 dimensional context feature vector for each event in this dataset. In contrast to the publicly available datasets, here there is no need for simulation of the known and unknown contexts; in a live setting the query features are immediately calculable or known, whereas the confidence and intent necessary to build a skill's feature set are unknown until a skill is executed. Because the confidence and intent for a skill are both accessible post execution, we reveal them together. We accommodate this by slightly modifying the objective of CATSO to reveal $U$ unknown skill feature sets instead of $U$ unknown individual features for each event. 

\begin{figure}[h]
\centering
\begin{minipage}{.49\textwidth}
  \centering
  \includegraphics[width=\linewidth]{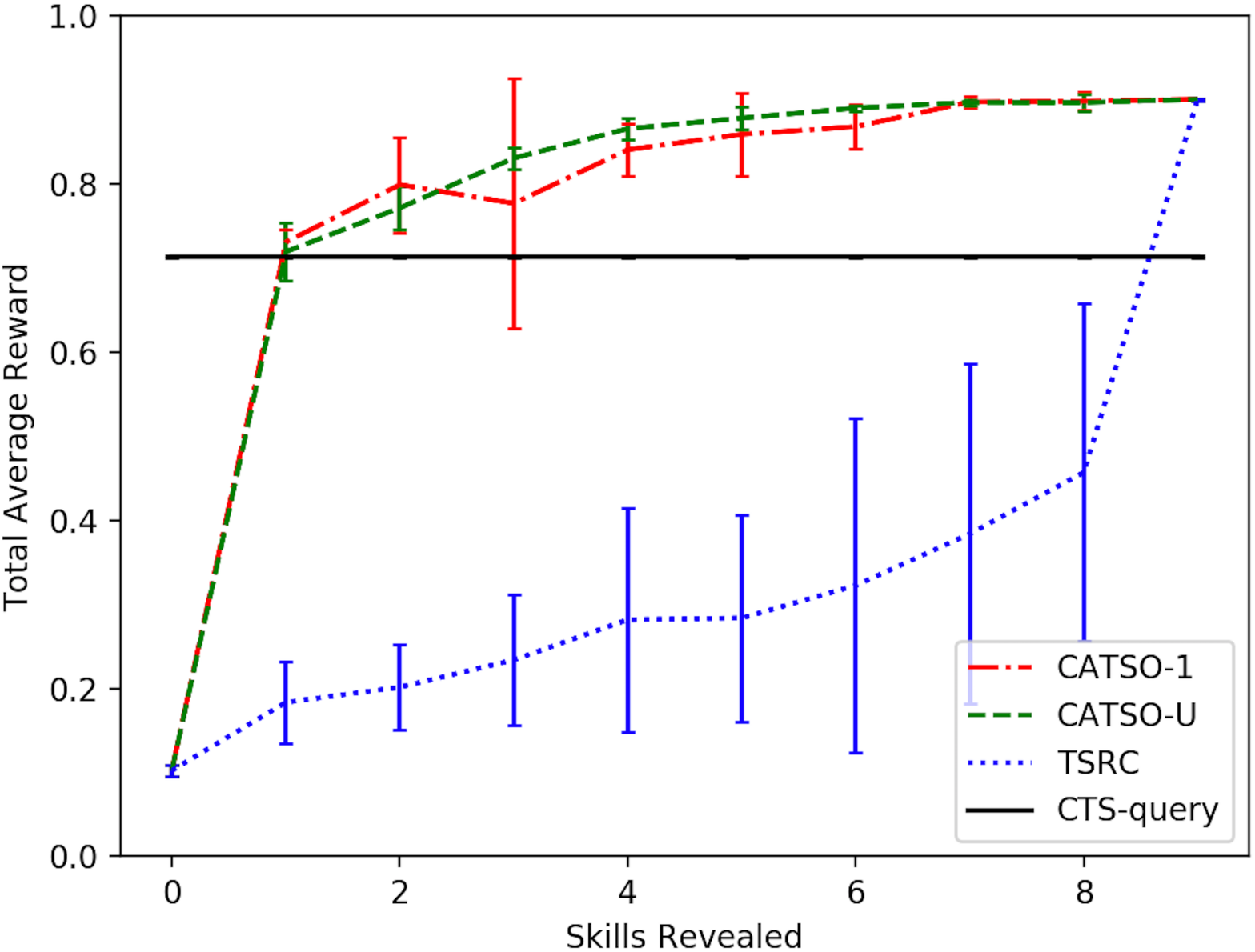}
  \captionof{figure}{Stationary Setting - Customer Assistant with 9 Skills}
  \label{fig:chipS}
\end{minipage}%
\hfill
\begin{minipage}{.49\textwidth}
  \centering
  \includegraphics[width=\linewidth]{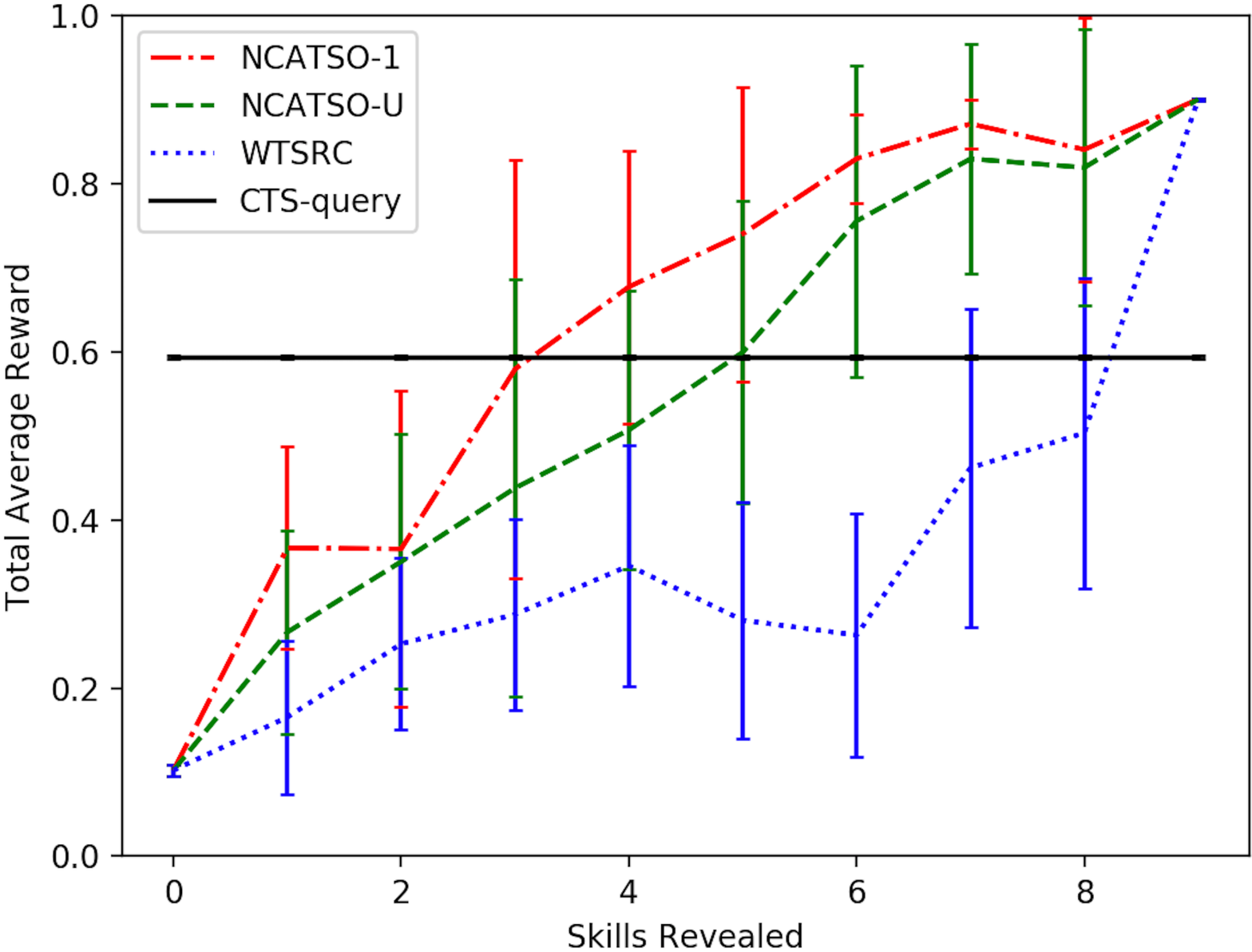}
  \captionof{figure}{Nonstationary Setting - Customer Assistant with 9 Skills}
  \label{fig:chipN}
\end{minipage}
\end{figure}

We perform a deeper analysis of the \textit{Customer Assistant} dataset, examining the case where $S=U$. Recall that when $S=1$, the known context, in this case the query features, is used to select all $U$ additional context features sets at once, whereas when $S=U$, the known context grows and is used to select each of the $U$ additional context feature sets incrementally. Maintaining $\lambda(t)=1$, for the stationary case we denote these two cases of the CATSO algorithm as CATSO-1 and CATSO-U respectively and report their performance across various $U$, the number of unknown skill feature sets revealed. Note that when all 9 skill feature sets are revealed, the CATSO and TSRC methods all reduce to simple Contextual Thompson Sampling (CTS) with the full feature set. Similarly, when 0 skill feature sets are revealed, the methods all reduce to CTS with a sparsely represented context of the query features. CTS suffers from this sparsity so we also consider a case we call CTS-query, CTS where the context is exclusively the query features. CTS-query is thus an a-priori online approach to dialog orchestration that completely ignores the existence of post-execution features. The results for the stationary case are summarized in Figure \ref{fig:chipS}. CATSO-U appears to slightly outperform CATSO-1 across all $U$ tested and both methods outperform TSRC by a large margin. Also notice that our posterior methods CATSO-1 and CATSO-U outperform the a-priori method CTS-query even under very small post-execution feature budgets, as low as 2 skill feature sets. 

For the nonstationary case we simulate nonstationarity in the same manner as the publicly available datasets, except using the natural partition of the query features as the known context and the skill feature sets as the unknown context instead of simulated percentages. We use the GP-UCB algorithm for $\lambda(t)$, refer to the $S=1$ and $S=U$ cases as NCATSO-1 and NCATSO-U, and illustrate their performance alongside WTSRC and CTS-query in Figure \ref{fig:chipN}. Here we observe that NCATSO-1 slightly outperforms NCATSO-U, and both outperform the WTSRC baseline. Notice that posterior approach NCATSO-1 outperforms CTS-query, the a-priori approach, when approximately 3 or more skill feature sets are revealed.  

\section{Conclusions and Future Work}
In this paper we consider how to address the challenges of posterior dialog orchestration using an online approach. We formulated CABO, a new variant of context attentive bandits motivated by practical budgets on skill execution and demonstrate that our new bandit algorithm beats the existing state of the art context attentive bandit algorithm on simulated (non-dialog) and dialog datasets across stationary and nonstationary settings.

\textit{Customer Assistant} has now been deployed to over 100,000 users with a thumbs up/down feature that allows users to provide individual feedback to each of the responses. The system also allows Subject Matter Experts (SMEs) to provide explicit labels to each event, enabling human evaluation of all the models. 

Theoretical regret bounds on the proposed algorithm will follow in a more theory focused work. Our current algorithm treats multi-skill dialog orchestration as a single class problem, where after each query, only one skill response is returned to the user. However, in many use cases, multiple responses ought to be returned to the user. We would like to shift our algorithm to the multiclass setting, perhaps by using the contextual combinatorial bandit approach in the arm selection process in addition to its current role in the unknown context feature selection process. Also, our current formulation assumes that all of the unobserved features are of the same cost, and thus the budget on cost is equivalent to a budget on the number of features. We plan on expanding this notion of budget to accommodate settings where unobserved features have different costs. Other directions for future work include using non-bandit algorithms in the context feature selection stage and exploring nonstationarity in the known context space.

\bibliographystyle{unsrt}
\bibliography{main}
\end{document}